\documentclass[10pt,twocolumn,letterpaper]{article}

\usepackage{wacv}
\usepackage{times}
\usepackage{epsfig}
\usepackage{graphicx}
\usepackage{amsmath}
\usepackage{amssymb}
\usepackage{booktabs}
\usepackage{multirow}
\usepackage{pifont}
\newcommand{\xmark}{\ding{55}}%

\def\wacvPaperID{1266} 

\wacvapplicationstrack 

\wacvfinalcopy 

\ifwacvfinal
\usepackage[breaklinks=true,bookmarks=false]{hyperref}
\else
\usepackage[pagebackref=true,breaklinks=true,colorlinks,bookmarks=false]{hyperref}
\fi

\begin{document}

\title{EmbryosFormer: Deformable Transformer and Collaborative Encoding-Decoding for Embryos Stage Development Classification}

\author{Tien-Phat Nguyen\textsuperscript{1,4}, Trong-Thang Pham\textsuperscript{*4}, Tri Nguyen\textsuperscript{*7}, Hieu Le\textsuperscript{*1}, Dung Nguyen\textsuperscript{5}, \\ Hau Lam\textsuperscript{6}, Phong Nguyen\textsuperscript{1}, Jennifer Fowler\textsuperscript{8}, Minh-Triet Tran\textsuperscript{2,3,4}, Ngan Le\textsuperscript{9} \\
\small \textsuperscript{1}FPT Software AI Center, Ho Chi Minh City, Vietnam \\
\small \textsuperscript{2}University of Science, VNU-HCM; 
\small \textsuperscript{3}Vietnam National University, Ho Chi Minh City, Vietnam\\
\small \textsuperscript{4}John von Neumann Institute, Vietnam National University, Ho Chi Minh City, Vietnam\\
\small \textsuperscript{5}IVFMD, My Duc Phu Nhuan hospital, Ho Chi Minh City, Vietnam\\
\small \textsuperscript{6}Olea Fertility, Vinmec Central Park International Hospital, Ho Chi Minh City, Vietnam\\
\small \textsuperscript{7}HOPE Research Center, My Duc Hospital, Ho Chi Minh City, Vietnam \\
\small \textsuperscript{8}Arkansas Economic Development Commission, Little Rock, AR USA 72202\\
\small \textsuperscript{9}Department of Computer Science and Computer Engineering, University of Arkansas, Fayetteville, AR, USA 72703\\
\small \textsuperscript{*}\textit{equal contribution}\\
}

\maketitle
\thispagestyle{empty}

\begin{abstract}
The timing of cell divisions in early embryos during the In-Vitro Fertilization (IVF) process is a key predictor of embryo viability. However, observing cell divisions in Time-Lapse Monitoring (TLM) is a time-consuming process and highly depends on experts. In this paper, we propose EmbryosFormer, a computational model to automatically detect and classify cell divisions from original time-lapse images. Our proposed network is designed as an encoder-decoder deformable transformer with collaborative heads. The transformer contracting path predicts per-image labels and is optimized by a classification head. The transformer expanding path models the temporal coherency between embryo images to ensure monotonic non-decreasing constraint and is optimized by a segmentation head. Both contracting and expanding paths are synergetically learned by a collaboration head. We have benchmarked our proposed EmbryosFormer on two datasets: a public dataset with mouse embryos with 8-cell stage and an in-house dataset with human embryos with 4-cell stage. Source code: \url{https://github.com/UARK-AICV/Embryos}.

\end{abstract}
\section{Introduction}
Fertility impairment affects approximately 80 million people globally, with one in every six couples undergoing infertility issues \cite{SART, minhas2021european}. This necessitates the use of IVF for conceiving. During the IVF procedure, a patient is stimulated to produce multiple oocytes. Then, a fraction of them fertilises, and a smaller fraction continues to grow and develop normally as embryos before being transferred into the uterus. Because of the increased maternal and fetal risks associated with multi-fetal gestation, only one embryo with the highest viability should be chosen for implantation at a time \cite{norwitz2005maternal, practice2017guidance}. Clinically, embryologists select potential embryos manually by considering the morphological features and rate of development. Unlike the traditional monitoring process, where embryos are taken out of the incubators at discrete time points, Time-Lapse Monitoring (TLM) techniques offer more general and uninterrupted observations on embryo development processes, where embryos are being kept safely in their culprits without any external intervention while built-in microscope systems periodically capture data of the embryos inside \cite{nakahara2010evaluation}. However, TLM still requires human expertise and experience. Consequently, results often come with variability and large labor expenses. Therefore, there is an emerging demand for developing an automated and time-effective tool to support embryologists in the selection processes.

Embryo morphology is captured at discrete time points in real-world settings. As a result, the characteristics or position of the embryo can vary rapidly and unexpectedly from frame to frame. Recently, Deep Neural Networks (DNNs), particular Convolutional Neural Networks (CNNs) have made significant progress in providing decision-making solutions at the human-expert level. Their successes have been reported in different fields and modalities among the diagnostic medical domain \cite{shen2017deep, zhou2021deep}, such as chest x-rays abnormalities recognition \cite{guendel2018learning, le2021pairflow, le2021enhance}, provision of biomarkers of tumors on MRI images \cite{ho2021point, nguyen20213d, yamazaki2021invertible}, organs structure analysis on MRI images \cite{le2021offset, le2021multi, hoang2022dam, tran2022ss}. DNNs have been recently applied to the task of classifying embryo stage development. Existing methods \cite{leahy2020automated, lockhart2021automating, lukyanenko2021developmental,malmsten2021automated, mcauley2018predicting} consider time-lapse embryos videos as sequences of images and utilize 2D-CNNs to perform per-frame classification, then apply a post-processing step with dynamic programming to enforce the predictions following the monotonic non-decreasing constraint. Such approaches deal with the high imbalance across classes as well as can not consider temporal information.
Other works \cite{lukyanenko2021developmental, lockhart2021automating} introduce two-stream networks that incorporate temporal information to address the imbalance issue while incorporating the monotonic constraint into the learning stage. Despite showing promising results, these methods process a fixed size of frame sequence at a time, which could lack the global context of the entire video and also increase the inference time. In this work, we utilize deformable Transformer \cite{zhu2020deformable} to propose an encoder-decoder deformable Transformer network for embryos stage development classification. Our proposed network contains three heads aiming classification, segmentation, and refinement.
Our contribution is two-fold as follows:

\noindent
$\bullet$ \textbf{Dataset}: We have conducted an Embryos Human dataset with a total of 440 time-lapse videos of 148,918 images, gathered from a real-world environment and collected from a diverse number of patients. The dataset has been carefully pre-processed, annotated and conducted by three embryologists. The data will be made available for the research community, please contact the author.

\noindent 
$\bullet$ \textbf{Methodology}: We propose EmbryosFormer, an effective framework for monitoring embryo stage development. Our network is built based on the Unet-like architecture with deformable transformer blocks and contains two paths. 
A contrasting path (i.e. deformable transformer encoder) aims to predict per-class label, whereas an expanding path (i.e. deformable transformer decoder) models stage-level by taking temporal consistency into consideration. The feature encoding at the encoding path is optimized by a classification head, and the temporal coherency at the decoding path is trained by a segmentation head. Both encoding and decoding paths are cooperatively learned by a collaboration head. We empirically validate the effectiveness of our proposed EmbryosFormer by showing that, to the best of our knowledge, it achieves superior performance to \textit{all} of the current state-of-the-art methods benchmarked on the two datasets of Embryos Mouse and Human.

\section{Related Work}
\subsection{Detection Transformer}
\label{sec:DETR}
The core idea behind transformer architecture \cite{vaswani2017attention} is the self-attention mechanism to capture long-range relationships. Transformer has been successfully applied to enrich global information in computer vision \cite{yang2020learning, chen2021pre, vo2021aei, touvron2021training}. When it comes to object detection, Detection Transformer (DETR) \cite{carion2020end} is one of the most well-known approaches, which performs the task as a set prediction. Unlike traditional CNNs-based methods \cite{ren2015faster,he2017mask}, Detection Transformer (DETR) \cite{carion2020end} performs the task as a set prediction. Even DETR obtains good performance while providing an efficient way to represent each detected element, it suffers from high computing complexity of quadratic growth with image size and slow convergence of global attention mechanism. The recent Deformable Transformer \cite{zhu2020deformable} is proposed to address the limitations while gaining better performance by incorporating multi-scale feature representation and attending to sparse spatial locations of images. Not only in the image domain, but DETR is also successfully applied to the video domain e.g. dense video captioning PDVC \cite{wang2021end}. 

\subsection{Embryo stage development classification}
\label{sec:rw_stagedev}
Classifying embryo development stages aims to provide a cue for quality assessment of fertilized blastocysts, which requires complex analyses of time-lapse imaging videos besides identifying development stages.
Traditionally, embryologists must review the embryo images to determine the time of division for each cell stage development. This process does require not only expert knowledge but also experience and is time-consuming. With the emergence of DNNs, CNNs have been used to assess embryo images. Generally, DNNs-based  embryo stage development classification can be divided into two categories: image-based and sequence-based. In the first group, Khan et al., \cite{khan2016deep} utilizes CNNs (i.e., AlexNet) \cite{krizhevsky2012imagenet} and a Conditional Random Field (CRF) \cite{sutton2012introduction} to count human embryonic cell over the first five cell stages. Ng et al., \cite{ng2018predicting} used ResNet \cite{he2016deep} coupled with a dynamic programming algorithm for post-processing to predict morphokinetic annotations in human embryos. Later, Lau et al., extend \cite{ng2018predicting} with region of interest (ROI) detection and LSTM \cite{gers2000learning} for sequential classification. Rad et al., \cite{rad2019cell} proposes Cell-Net, which uses ResNet-50 \cite{he2016deep} to parse centroids of each cell from embryo image. Leahy et al., \cite{leahy2020automated} extracts five key features from time-lapse videos, including stage classification, which utilizes ResNeXt101 \cite{Xie2016} to predict per-class probability for each image. Malmsten et al., \cite{malmsten2021automated} uses Inception-V3 \cite{szegedy2016rethinking} to classify human embryo images into different cell division stages, up to eight cells. While showing promising results on automatically classifying embryonic cell stage development with DNNs, image-based prediction approaches ignore temporal coherence between time-lapse images and the monotonic development order constraint during training. In the second group, Lukyanenko et al., \cite{lukyanenko2021developmental} incorporate CRFs \cite{sutton2012introduction} to include the monotonic condition into the learning process for sequential stage prediction. Lockhart et al., \cite{lockhart2021automating} propose synergic loss and utilize LSTM \cite{gers2000learning} and VGG-16 \cite{simonyan2014very} to enable the network recognize and utilize stage similarities between different embryos. Malmsten et al.,\cite{malmsten2021automated} propose CellDivision, which uses InceptionV3 \cite{szegedy2015going} to extract visual encoding feature. In CellDivision, misclassification is handled by a post-processing step of global optimization. 

Our network belongs to the second group. Inspired by the success of DETR \cite{carion2020end} we extend Deformable Transformer \cite{zhu2020deformable} with a collaborative encoding-decoding to improve time-lapse embryos stage development classification. 


\begin{figure*}[t!]
\centering
\includegraphics[width=0.85\textwidth]{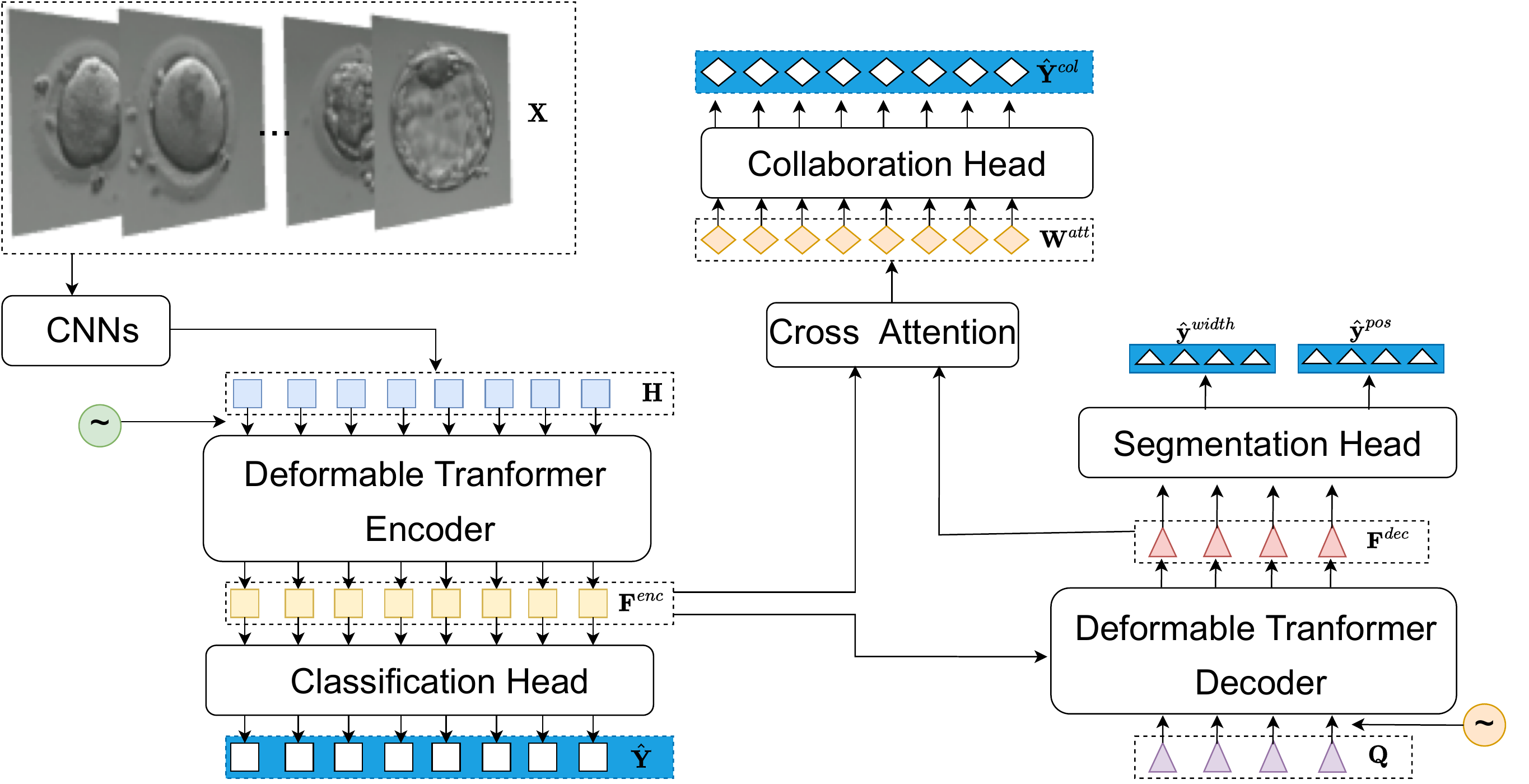}
\caption{The overall flowchart of our proposed EmbryosFormer for the embryo stage development classification. EmbryosFormer is a framework with three heads: classification head, segmentation head and collaboration head. At the feature encoding path, the image-level feature $\mathbf{H}$ is first extracted by adopting a 2D CNNs into the time-lapse embryos images $\mathbf{X}$. The encoding feature $\mathbf{F}^{enc} \in \mathbb{R}^{T\times C\times d}$ is then obtained by applying a deformable transformer encoder into image-level feature $\mathbf{H}$. The feature encoding is trained on classification loss. At the decoding path, a deformable transformer decoder is adopted, in which stage query $\mathbf{Q}$ and encoding feature $\mathbf{F}^{enc}$ serve as query and key, respectively, to obtain $\mathbf{F}^{dec} \in \mathbb{R}^{C\times d}$. The decoding is trained on segmentation loss. Both encoding feature $\mathbf{F}^{enc}$ and decoded query feature $\mathbf{F}^{dec}$ are then fed to a cross-attention to produce an attention weight $\mathbf{W}^{Att} \in \mathbb{R}^{T\times C}$ which 
cooperatively trained by a collaboration head.}
\label{fig:overall}
\end{figure*}

\section{The Proposed Method}
Our proposed network is illustrated in Fig. \ref{fig:overall}, which contains two paths corresponding to contracting and expanding. The contracting path aims to encode the visual representation of each time-lapse embryos images, optimized by a classification head. The contracting network is designed as a deformable transformer encoder. The expanding path models temporal consistency to ensure monotonic constraint between time-lapse embryos by processing the stage classification as a segmentation task. The expanding network is designed as a deformable transformer decoder and optimized by segmentation loss. Both encoder and decoder networks are synergistically learned in a framework with three heads: classification head, segmentation head and collaboration head. 

\subsection{Problem Setup}
\label{sec:setup}
Given a time-lapse embryo video $\mathbf{X}$, which is considered as a sequence of embryo images, i.e. $\mathbf{X} = [\mathbf{x}_1, \mathbf{x}_2, ..., \mathbf{x}_T]$, where $T$ is the sequence length, our objective is to predict a sequence of image-wise stage labels $\mathbf{\hat{Y}} = [\hat{\mathbf{y}}_1, \hat{\mathbf{y}}_2, ..., \hat{\mathbf{y}}_T]$, where $\hat{\mathbf{y}}_i \in \mathbb{R}^C$, and $C$ is the number of stages. During training, each embryos image $\mathbf{x}_i$ is associated with a label $\mathbf{y}_i \in \mathbb{R}^C$ and the time-lapse embryo video $\mathbf{X}$ has its corresponding groundtruth label $\mathbf{Y} = [\mathbf{y}_1, \mathbf{y}_2, ..., \mathbf{y}_T]$. The label is constrained by monotonic non-decreasing, thus, the groundtruth label $\mathbf{Y}$ is also presented by a set of segments $\mathbf{Y} =\{s_c\}|_{c=1}^C$, where $s_c = [\mathbf{x}_k, \mathbf{x}_{k+1}, ..., \mathbf{x}_{k+t}]$ with $t$ is the number of images in the $c^{th}$ segment $s_c$, and $\mathbf{\hat{Y}} =\{\hat{s}_c\}|_{c=1}^C$, where $\hat{s}_c = [\mathbf{x}_{k'}, \mathbf{x}_{k'+1}, ..., \mathbf{x}_{k'+t'}]$ with $t'$ is the number of images in the predicted segment $\hat{s}_c$. Each embryo segment is presented by its width and center position.

\subsection{Deformable Transformer: A Revise}
  Given a multi-scale input feature map $\mathbb{F} =\{\mathbf{f}^l\}|_{l=1}^{L}$, where $\mathbf{f}^l \in \mathbb{R}^{C_l \times H_l \times W_l}$, the $K$ sampling locations of interest are generated for each feature map at $l^{th}$ level, and $a^{th}$ attention head. For a query $\mathbf{q}_i$, coordinates of sampling locations are estimated through reference points $p_i \in [0, 1]^2$ and $K$ sampling offset $\Delta p$. Corresponding to the $a^{th}$ attention head at the $l^{th}$ feature map $\mathbf{f}^l$ and the $i^{th}$ query element $\mathbf{q}_i$,  the $k^{th}$ sampling location is defined by
\begin{equation}
    \hat{p}_{alik} = \phi_l(.) + \Delta p_{alik}
\end{equation}, where  $\phi_l(.)$ is a function to project normalized reference points to the input feature map $\mathbf{f}^l$. $\Delta p_{alik}$ is sampling offset w.r.t. $\phi_l(.)$. Multi-scale deformable attention is defined as follows:
\begin{equation}
    f(q_i, p_i, \mathbb{X}) = \sum_{a=1}^{N_a} W_a\sum_{l=1}^L\sum_{k=1}^K A_{alik}\mathbf{W}_a\textbf{x}^l_{\hat{p}_{alik}}
\label{eq:def_t}
\end{equation}
where $a, l, k$ are index of the attention head, input feature level, the sampling location, respectively. $A_{alik}$ denotes an attention weight for the $a^{th}$ attention head, the $l^{th}$ feature level, the $i^{th}$ query element and the $k^{th}$ sampling location. $\mathbf{W}_a$ is the projection matrix for key elements. 

\subsection{Feature Encoding}
\label{sec:enc}
The feature encoding network takes a sequence of time-lapse embryo images $X$ as input and output encoded feature $\mathbf{F}^{enc}$. Different from videos, embryo morphology is captured at discrete times. Namely, time-lapse embryo images have been collected by an incubator system consisting of a built-in camera and a microscope, which takes an image of embryos every particular time. Thus, well-known video feature encoding methods such as C3D, and I3D may not be appropriate approaches because blastocyst characteristics of nearby embryo images are much different. In this work, we first adopt a pre-trained 2D CNNs model (Resnet50 is used as an instance) to extract visual feature $\mathbf{h}_i \in \mathbb{R}^{2048}$ of each individual embryo image $\mathbf{x}_i$. As a result, the time-lapse embryos images $\mathbf{X} = \{\mathbf{x}_i\}|_{i=1}^T$ is represented by $\mathbf{H} = \{\mathbf{h}_i\}|_{i=1}^T$. 
To enrich the representation with multi-scale features, we process the feature map with temporal convolutional layers of stride as 2. To extract the inter-image coherence across multiple scales, we then encode the multi-scale embryo features with their positional embedding \cite{vaswani2017attention} into the deformable self-attention \cite{zhu2020deformable}. Specifically, we apply multi-scale deformable attention defined in Eq.\ref{eq:def_t} into feature $\mathbf{H}$. As a result, we obtain encoding feature $\mathbf{F}^{enc} \in \mathbb{R}^{T \times C\times d^{enc}}$, where $d^{enc}$ is the encoder embedding dimension and $C$ is the number of stages. The feature encoding network extracts the visual representation of each individual embryo and its corresponding image-level label by a classification head. However, the monotonic non-decreasing constraint of time-lapse embryo images is not taken into consideration. More details of classification head is given section \ref{sec:colla}.

\subsection{Segmentation Decoding}
\label{sec:dec}
The decoding network considers temporal consistency to address the monotonic non-decreasing constraint by processing the embryo stages development classification as an embryo stages segmentation task. 
In this setup, each monotonic stage is considered as one segment and conditioned on a learnable embedding query $q_c \in \mathbb{R}^{d^{dec}}$, $c \in [1, C]$, where $C$ is the number of cell stages. Specifically, each query is assigned to a specific stage center. A positional embedding \cite{vaswani2017attention} is added to the decoder input to inject label information into each query. In this setup, the decoding network contains a stack of deformable cross attention \cite{zhu2020deformable} layers, in which the encoding feature $F^{enc}$ serves as keys and stage queries are defined as $ \mathbf{Q} = \{ q_c\}|_{c=1}^{C} \in \mathbb{R}^{C \times d^{dec}}$. Adopted by \cite{zhu2020deformable}, the stage queries $\mathbf{Q}$ serve as an initial guess of the embryos' feature, and they will be refined iteratively at each decoding layer. The output of query feature is denoted as $\mathbf{F}^{dec} \in \mathbb{R}^{C \times d^{dec}}$ and is optimized by a segmentation head, which is detailed in section \ref{sec:colla}.

For a simple setup, we set the same hidden size for both encoding and decoding paths, $d^{dec} = d^{enc} = d$. That is to say the encoding feature from the deformable transformer encoder is $\mathbf{F}^{enc} \in \mathbb{R}^{T\times C \times d}$ and decoded query feature from deformable transformer decoder $\mathbf{F}^{dec} \in \mathbb{R}^{C \times d}$.

\subsection{Collaborative Learning of Multiple Head}
Our proposed EmbryosFormer is designed as an Unet-liked framework and learned by three heads: classification head to predict the image-level label, segmentation head to simultaneously handle the monotonic constraint and predict stage-level label, collaboration head to strengthen image-level label with monotonic non-decreasing constraint. The overall flowchart of EmbryosFormer is given in Fig.\ref{fig:overall}.

\vspace{1mm}
\noindent
\textbf{Classification head:} Although temporal information contains valuable information for stage detection, independent embryo images also contain specific morphological features that characterize its stages. Hence, an image-wise classification head is added to the feature encoding path. The classification head is learned by an image-level cross-entropy loss. Given a time-lapse embryos images $\mathbf{X} = \{\mathbf{x}_i\}|_{i=1}^T$, where $T$ is the number of images, the predicted image-level labels $\hat{\mathbf{Y}} = \{\hat{\mathbf{y}}_{i}\}|_{i=1}^T \in \mathbb{R}^{T\times C}$ is computed by applying a feed-forward network (FFN) then a softmax function into encoded feature $\mathbf{F}^{enc}$ as below:
    \begin{equation}
        \mathbf{\hat{Y}} = \mathrm{Softmax}(\mathrm{FFN}(\mathbf{F}^{enc}))   
    \end{equation}
The classification head is learned by an image-level cross entropy function as follows:
\begin{equation}
    \mathcal{L}^{cls} = \sum_{i=1}^{T} -\mathbf{y}_i \log{\hat{\mathbf{y}}_i}
\end{equation}

\noindent
\textbf{Segmentation head:}
The decoded query feature $\mathbf{F}^{dec}$ aims to predict stage-level labels by taking temporal coherence of monotonic constraint into account. Each $c^{th}$ stage development is assigned one segment $s_c = \{\mathbf{x}_k, \mathbf{x}_{k+1}, ..., \mathbf{x}_{k+t}\}$, which is represented by a segmentation width ${y}_c^{width} = t$ frame and a center position  $\mathrm{y}_c^{pos} = k+t/2$. The segmentation head contains a softmax layer after an FFN to normalize and turn the stage-level predictions into unit fractions. Hence, to enforce the predictions following the monotonic non-decreasing constraint, the decoding network partitions the input sequence into a list of stage-level segments without the need for any post-processing techniques. The predicted segmentation width and center position of all $C$ stage-level segments are defined as follows:

\begin{equation}
        \hat{\mathbf{y}}^{width} = \mathrm{Softmax}(\mathrm{FFN}(\mathbf{F}^{dec}))   
    \end{equation}
    Stage center positions are calculated based on the cumulative sum of the predicted stage widths:
    \begin{equation}
        \hat{\mathbf{y}}^{pos} = \mathrm{CumulativeSum}(\hat{\mathbf{y}}^{width}) - \hat{\mathbf{y}}^{width}/2
    \end{equation}
\noindent
where $\hat{\mathbf{y}}^{width}, \hat{\mathbf{y}}^{pos} \in \mathbb{R}^C$. \\
The segmentation head is learned by L1 loss as follows:
\begin{equation}
    \mathcal{L}^{seg} = \|\mathbf{y}^{width} - \hat{\mathbf{y}}^{width}\|_1 + 
    \|\mathbf{y}^{pos} - \hat{\mathbf{y}}^{pos}\|_1
\end{equation}

\noindent
\textbf{Collaboration head:} This head utilizes stage-level decoded query embedding $\mathbf{F}^{dec}$ containing global information as a guided context to refine the image-level features $\mathbf{F}^{enc}$. We make use of a cross attention mechanism \cite{vaswani2017attention} to align image-level prediction and its corresponding stage-level prediction. Concretely, the image-level embeddings $\mathbf{F}^{enc} \in \mathbb{R}^{T \times C \times d} $ serve as queries, stage-level embeddings $\mathbf{F}^{dec} \in \mathbb{R}^{C\times d}$ serve as key elements, cross-attention is adopted to obtain attention weights $\mathbf{W}_{attn} = \{\mathbf{w}_i^{attn}\}|_{i=1}^T$, where $\mathbf{w}_i^{attn} \in \mathbb{R}^C$. The final frame-level classification $\hat{\mathbf{Y}}^{col} = \{\hat{\mathbf{y}}^{col}_i\}|_{i=1}^T$ is predicted as below:

\begin{equation}
        \hat{\mathbf{y}}_i^{col} = \mathrm{Softmax}(\mathbf{w}_i^{pos} \odot \mathbf{w}_i^{attn})
\end{equation}
where $\odot$ is the element-wise multiplication. The position weight $\mathbf{w}_i^{pos} \in \mathbb{R}^{C}$ is computed from the relative distance of the $i^{th}$ frame and the predicted stage segments. Concretely, given predicted stage segment width $\hat{\mathbf{y}}^{width} \in \mathbb{R}^C$ and segment position $\hat{\mathbf{y}}^{pos} \in \mathbb{R}^C$, the position weight $w_{i,c}^{pos}$ of 
the $i^{th}$ frame w.r.t $c^{th}$ stage is computed as below:
    \begin{equation}
        w_{i,c}^{pos} = \frac{\hat{y}_c^{width} +\alpha}{\left | \hat{y}_c^{pos} - i \right | + \alpha}
    \end{equation}
where $\alpha$ is a temperature constant monitoring the weight distribution. The collaboration loss is computed as a cross-entropy between the collaborative predicted label $\hat{\mathbf{Y}}^{col}$ and the groundtruth label $\hat{\mathbf{Y}}$
 \begin{equation}
    \mathcal{L}^{col} = \sum_{i=1}^{T} -\mathbf{y}_i \log{\hat{\mathbf{y}}^{col}_i}
 \end{equation}

The proposed EmbryosFormer is trained by three heads with loss $\mathcal{L}$ defined as follows:
\begin{equation}
\mathcal{L} = \mathcal{L}^{cls} + \mathcal{L}^{seg} + \mathcal{L}^{col}
\end{equation}

\label{sec:colla}
\section{Experimental results}
\subsection{Dataset}
\noindent
\textbf{Mouse Embryos Dataset} We use the NYU Mouse Embryo dataset \cite{cicconet2014label}, which contains 100 videos of developing mouse embryos. Embryo dataset \cite{cicconet2014label}, which contains 100 sequences of developing mouse embryos. Each time-lapse embryo imaging is constructed from a sequence of 480 $\times$ 480 images, captured every seven seconds. On average, each sequence consists of 314 images per embryo and 8 developmental stages. We follow \cite{lukyanenko2021developmental} to randomly divide the data into 80/10/10 for training, validating, and testing.

\begin{figure*}[t!]
\centering
\includegraphics[width=\textwidth]{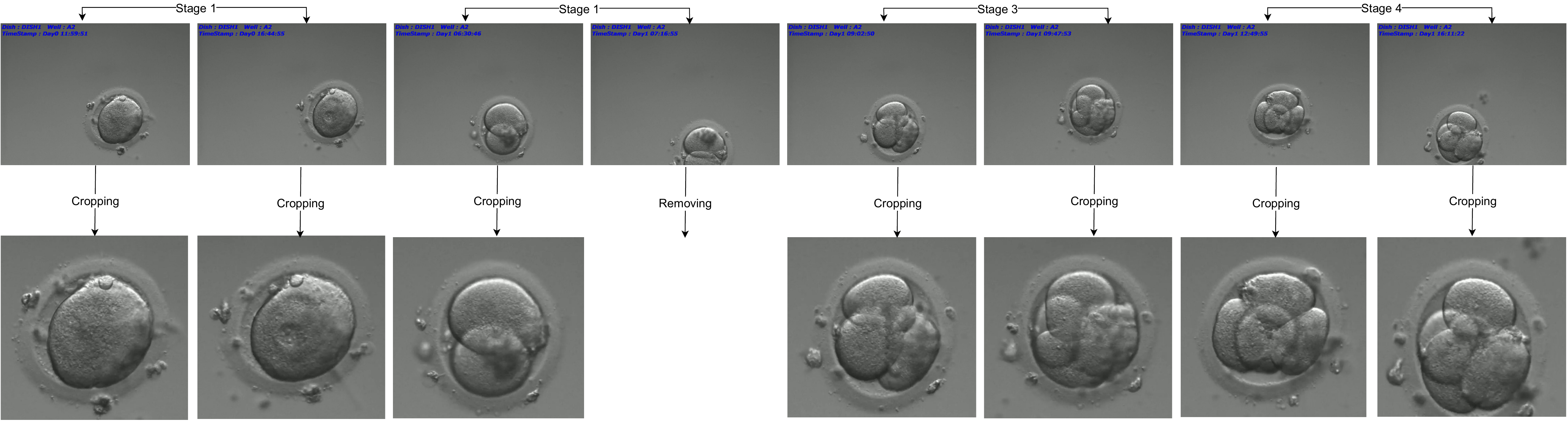}
\caption{Examples of developing human embryos at four-cell stages with capture time on the top of each image. Top: Original images from incubator system. Bottom: Embryos images after pre-processing (i.e. cropping and removing).}
\label{fig:examples}
\end{figure*}

\noindent
\textbf{Human Embryos Dataset}
The dataset consists of 440 time-lapse embryos of intracytoplasmic sperm injection (ICSI) and was collected from 112 patients with an average age of female and male are 33.0 $\pm $ 4.40 and 38.2 $\pm $ 7.47 years, respectively.
Furthermore, the incubator system ASTEC CCM-iBIS captures images while shifting among multiple dishes in the unit. Therefore, all images within a time-lapse embryo are not aligned well and contain redundant backgrounds, as shown in Fig.\ref{fig:examples} (top). 
We pre-processed the data by removing all images not covering the embryo's cells and cropping the redundant boundary as shown in Fig.\ref{fig:examples} (bottom). 
After pre-processing, we received time-lapse embryos with images of size about $400 \times 400$ captured at least every fifteen minutes. On average, each time-lapse embryos images consists of 339 images and 4 developmental stages. The Human Embryos dataset was pre-processed and annotated by three embryologists.

We randomly divide each dataset into 80/10/10 for training validating, and testing. Table \ref{tb:data} shows the cell stage distribution for all images over Embryos Mouse and Embryos Human datasets.

\begin{table*}[]
\centering
\caption{Data distribution of each cell stage in Embryos Mouse and Embryos Human datasets.}
\resizebox{\textwidth}{!}{
\begin{tabular}{l|rrrrrrrr|r|rrrr|r}
\toprule
      \textbf{Cell} & \multicolumn{8}{c|}{\textbf{Mouse}} & \multirow{2}{*}{\textbf{Total}}& \multicolumn{4}{|c|}{\textbf{Human}} &  \multirow{2}{*}{\textbf{Total}} \\ \cline{2-9} \cline{11-14}
      \textbf{Stage} & 1     & 2 & 3 & 4 & 5 & 6 & 7 & 8 & & 1     & 2 & 3 & 4 &\\ \toprule
Train & 4,033  & 13,308  & 654  & 7,053  & 377  & 651  & 682  & 160  & 26,918& 22,393 &  57,434 & 14,581  & 24,132 & 118,540\\ 
Val   &   452   & 1,626  & 62  & 863  & 46  & 34  & 57  &  20 & 3,160 &  2,592  & 8,047  & 1,775  & 2,937 & 15,351\\
Test  & 517  & 1,598  & 112  & 904  & 58  & 120  & 74  & 20  & 3,403 & 2,978 & 7,006  & 1,835 & 3,208 & 15,027\\ \hline
\textbf{Total} & 5,002  & 16,532  & 828  & 8,820  & 481  & 805  & 813  & 200  &  & 27,963 & 72,487 & 18,191  & 30,277 & \\ \bottomrule
\end{tabular}}
\label{tb:data}
\end{table*}

\subsection{Metrics}
We use three common metrics, i.e, Precision, Recall and Accuracy, to evaluate the proposed method at the global level of the entire video and per-class level as follows:

\noindent
\textbf{Global accuracy}: per-instance accuracy, the number of true predictions over the total number of embryo images.

\noindent
\textbf{Per-class precision/recall}: As some stages dominate large parts of the embryo development process, we compute per-class precision and recall, then take their average values for general comparison.

\subsection{Implementation Details}
 
We follow \cite{lukyanenko2021developmental}, each dataset is divided into 80/10/10 for training/validation/testing. To conduct a fair comparison with SOTA existing methods, we benchmark our EmbryosFormer with ResNet-50 \cite{he2016deep} as a CNNs feature encoder which was pre-trained on ImageNet \cite{krizhevsky2012imagenet}. 
We use a two-layer deformable transformer with two levels of multiscale deformable attention. The deformable transformer uses a hidden size of 512 and 512 in feed-forward layers. 
The whole network is trained with a batch size of 32 and 250 epochs.
We use Adam \cite{kingma2014adam} optimizer, $10^{-4}$ weight decay with linear warm-up and cosine annealing decay scheduler (i.e. learning rate follows a linear warmup schedule between $0$ and $10^{-3}$ in the first $20$ epochs followed by a cosine annealing schedule between $10^{-3}$ and $0$).

\subsection{Performance Comparison}
In this section, we conduct the performance comparison between our proposed EmbryosFormer with the existing SOTA approaches including CNNs-CRF \cite{lukyanenko2021developmental}, ESOD \cite{lockhart2021automating}, and CellDivision \cite{malmsten2021automated}. To further investigate the effectiveness of our proposed approach, we benchmark EmbryosFormer with different configurations as below:

\noindent
$\bullet$ \textit{CNNs (baseline)}: In this baseline setup, the visual feature of each embryos image is extracted by ResNet-50 \cite{he2016deep} and the classification is conducted by a softmax function.

\noindent
$\bullet$ \textit{CNNs-Trans}: In this network configuration, the visual encoding feature of each embryos image is extracted by ResNet50 \cite{he2016deep}. The temporal relationship between images is modelled by Transformer
    \cite{vaswani2017attention}. 
    
\noindent
$\bullet$ \textit{CNNs-Trans-CRF}: Similar to the previous network i.e., CNNs-Trans, in which the visual feature is extracted by  ResNet50 \cite{he2016deep} and the temporal relationship is model by Transformer. In this network setup, we further include the embryo monotonic constraint, following CNNs-CRF \cite{lukyanenko2021developmental}.

\noindent
$\bullet$ \textit{The proposed EmbryosFormer}: To fairly compare with the existing work, the visual feature of each embryo images is extracted by ResNet-50 \cite{he2016deep}. Our proposed EmbryosFormer is equipped with an encoder-decoder deformable transformer and trained by collaborative learning of three heads.

\vspace{1mm}
\begin{table*}[!ht]
    \centering
    \caption{Overall performance on global accuracy (Global Acc.), the average of per-class precision (Avg Pre.) and recall scores (Avg Rec.) on both Embryos Mouse \cite{cicconet2014label} and our in-house Embryos Human datasets. All scores are reported in \%. The best scores are highlighted in \textbf{bold}.}
    \resizebox{0.9\textwidth}{!}{
    \begin{tabular}{c|lll|ccc|ccc}
    \toprule
    & \multirow{3}{*}{\textbf{Method}}& \multirow{3}{*}{\textbf{Venue}} &\multirow{3}{*}{\textbf{Feature}} & \multicolumn{3}{c|}{\textbf{Embryos Mouse}} & \multicolumn{3}{c}{\textbf{Embryos Human}} \\ \cline{5-10}
    &  & & & \shortstack{Global\\Acc.$\uparrow$} & \shortstack{Avg\\Pre$\uparrow$.} & \shortstack{Avg\\Rec$\uparrow$.} & \shortstack{Global\\Acc$\uparrow$.} & \shortstack{Avg\\Pre$\uparrow$.} & \shortstack{Avg\\Rec$\uparrow$.} \\ 
    \toprule
   &  CNNs-CRF \cite{lukyanenko2021developmental} & MICCAI`21 & ResNet-50 & 97.3& 91.8 & 81.1 & 88.6 & 86.8 & 82.2  \\ 
   & CellDivision \cite{malmsten2021automated} & Neu.Com.\&App`21 & InceptionV3 & 97.8  & 92.8  & 85.2 & 89.2 & 86.3 & 87.1 \\
   & ESOD \cite{lockhart2021automating} & MICCAI`21 & VGG-16 & 97.5  & 80.3 & 76.0  & 91.6 & 89.1 & 91.1 \\ 
    \midrule
   \multirow{4}{*}{\rotatebox{90}{Ours}}  & CNNs (baseline)& & ResNet-50  & 96.8 & 84.2 & 82.3 & 90.1 & 87.6 & 89.2 \\
    & CNNs-Trans & & ResNet-50 & 97.6 & 93.2 & 82.6 & 91.5 & 86.0 & 81.1 \\
    & CNNs-Trans-CRF & & ResNet-50 & 97.7 & 91.4 & 84.7 & 91.8 & 88.7 & 87.5 \\
    & \textbf{EmbryosFormer} & & \textbf{ResNet-50} & \textbf{98.4}  & \textbf{95.3}  & \textbf{90.2} & \textbf{94.3} & \textbf{92.9} & \textbf{92.4} \\ 
    \bottomrule
    \end{tabular}}
    \label{tab:global}
\end{table*}

\begin{table*}[!ht]
\centering
\caption{Per-class precision performance comparison on Embryos Mouse dataset \cite{cicconet2014label}. All scores are reported in \%. The best scores are highlighted in \textbf{bold} and the second best scores are \underline{underlined}.}
\resizebox{\textwidth}{!}{
\begin{tabular}{c|ll|cccccccc}
    \toprule
    & \multirow{2}{*}{\textbf{Method}} & \multirow{2}{*}{\textbf{Feature}} &  \multicolumn{8}{c}{\textbf{Precision}$\uparrow$ / \textbf{Recall}$\uparrow$}\\ \cline{4-11}
    &  & & \textbf{1} & \textbf{2} & \textbf{3} & \textbf{4} & \textbf{5} & \textbf{6} & \textbf{7} & \textbf{8} \\
    \toprule
    & CNNs-CRF \cite{lukyanenko2021developmental}  & ResNet-50 & \underline{99.6} / \underline{99.6} & \underline{99.8} / 99.8 & 97.3 / 82.6 & 98.3 / \textbf{100} & \textbf{100} / 54.9 & 67.6 / 54.7 & 84.4 / 43.4 & \underline{87.5} / 55.0 \\ 
    & CellDivision \cite{malmsten2021automated}  & InceptionV3 & \textbf{100} / \textbf{100} & 99.6 / \textbf{100} & 95.5 / 94.6 & \textbf{100} / 99.1 & 90.0 / 62.1 & 67.4 / 96.7 & \underline{89.8} / 59.5 & \textbf{100} / 70.0 \\
    & ESOD \cite{lockhart2021automating}  & VGG-16 & \underline{98.9} / \textbf{100} & 99.6 / 99.6 & \textbf{100} / 93.8 & \underline{98.9} / \textbf{100} & 94.1 / 27.6 & \underline{74.4} / \textbf{99.2} & 76.5 / \textbf{87.8} & 0.0 / 0.0 \\ 
    \midrule
     \multirow{4}{*}{\rotatebox{90}{Ours}} &  CNNs (baseline) & ResNet-50 & \underline{99.6} / \underline{99.6} & 99.5 / \underline{99.9} & 97.6 / \underline{97.3} & 96.3 / 99.2 & 52.6 / 48.3 & 72.7 / 80.0 & 89.5 / 48.7 & \underline{87.5} / \underline{85.0} \\
    & CNNs-Trans & ResNet-50 & \textbf{100} / \textbf{100} & \textbf{99.9} / \underline{99.9} & \textbf{100} / 88.4 & 98.5 / \underline{99.9} & 59.7 / 51.7 & \textbf{79.7} / \underline{97.5} & 77.3 / \underline{63.5} & \textbf{100} / 60.0 \\
    & CNNs-Trans-CRF & ResNet-50 & 99.4 / \underline{99.6} & \underline{99.8} / 99.8 & 97.7 / \underline{97.3} & 96.0 / \textbf{100} & 30.0 / \textbf{74.1} & 69.5 / 92.5 & 62.5 / 44.6	& 80.0 / 70.0  \\
    & \textbf{EmbryosFormer} & ResNet-50 & \textbf{100} / \underline{99.6} & \underline{99.8} / \underline{99.9} & \underline{99.1} / \textbf{98.2} & \textbf{100} / \textbf{100} & \underline{97.6} / \underline{69.0} & 71.8 / \underline{97.5}  & \textbf{93.9} / 62.2 & \textbf{100} / \textbf{95.0} \\
    \bottomrule
\end{tabular}}
\label{tab:precision-recall}
\end{table*}

Each of existing work CNNs-CRF \cite{lukyanenko2021developmental}, ESOD \cite{lockhart2021automating}, and CellDivision \cite{malmsten2021automated} have been evaluated on a random split. To fairly compare our proposed EmbryosFormer with other existing SOTA works, we have benchmarked EmbryosFormer and others on the same data split. All models have been trained until converged with 200 epochs, except ESOD needs 400-500 epochs.
The overall performance on global accuracy, precision and recall is shown in Table \ref{tab:global} whereas the detailed classification performance on per-stage precision and recall is provided in Table \ref{tab:precision-recall} for Embryos Mouse dataset \cite{cicconet2014label} and Tables \ref{tab:accuracy_human} for our Embryos Human dataset, respectively. At each table, the best performance is highlighted in \textbf{bold}. Our implementations on various network setups have shown that the performance by our \emph{CNNs-Trans} and \emph{CNNs-Trans-CRF} are slightly  better than CNNs-CRF\cite{lukyanenko2021developmental} and ESOD \cite{lockhart2021automating} on Embryos Mouse dataset. Compare to CNNs-CRF\cite{lukyanenko2021developmental} and CellDivision \cite{malmsten2021automated}, our \emph{CNNs-baseline} and \emph{CNNs-Trans} are compatible on Embryos Mouse dataset and are better on Human dataset. In general, network configurations of \emph{CNNs (baseline)}, \emph{CNNs-Trans}, and \emph{CNNs-Trans-CRF} are competitive with CNNs-CRF \cite{lukyanenko2021developmental}, CellDivision \cite{malmsten2021automated} on both datasets even CellDivision \cite{malmsten2021automated} is based on InceptionV3, which takes longer time to train and inference.  
ESOD \cite{lockhart2021automating} is based on a synergic comparison between nearby, similar stages. Thus it needs more time to converge. However, it models temporal coherency with LSTM, which is limited to vanishing gradient and sensitive to imbalanced classes. For instance, Embryos Mouse dataset contains only 160 images of stage \#8 and 13,308 images of stage \#2, ESOD performs poorly with 0.0\% precision and 0.0\% recall on stage \#8 while it reaches 99.6\% precision and 99.6\% recall on stage \#1 as shown in Table \ref{tab:precision-recall}. 

 Based on Table \ref{tab:global}, first,  it can be observed that the performance of our method, EmbryosFormer, stands out against other SOTA methods by a large margin of all metrics on both datasets. Table \ref{tab:precision-recall} further show the advantages of EmbryosFormer in handling imbalanced data. For instance, on Embryos Mouse dataset, the precision/recall gap between cell stage \#2 and cell stage \#8 are 12.3/44.8\%, 0.4/30.0\%, 99.6/99.6\% on CNNs-CRF \cite{lukyanenko2021developmental}, CellDivision \cite{malmsten2021automated} and ESOD \cite{lockhart2021automating} respectively whereas the gap is 0.2/3.2\% on EmbryosFormer.

\subsection{Ablation Studies}
\begin{table*}[!ht]
    \centering
    \caption{Per-class accuracy performance comparison on our Embryos Human dataset. All scores are reported in \%. The best scores are highlighted in \textbf{bold} and the second best scores are \underline{underlined}.}
    \resizebox{0.9\textwidth}{!}{
    \begin{tabular}{c|lll|cccc|cccc}
    \toprule
    & \multirow{2}{*}{\textbf{Method}} & \multirow{2}{*}{\textbf{Venue}} &\multirow{2}{*}{\textbf{Feature}}&  \multicolumn{4}{c}{\textbf{Precision}$\uparrow$}& \multicolumn{4}{c}{\textbf{Recall}$\uparrow$} \\ \cline{5-12}
    &  & & &  \textbf{1} & \textbf{2} & \textbf{3} & \textbf{4} & \textbf{1} & \textbf{2} & \textbf{3} & \textbf{4}\\
    \toprule
    & CNNs-CRF \cite{lukyanenko2021developmental} & MICCAI`21 & ResNet-50 & 90.4 & 90.1 & 71.2 & 95.6 & 87.1  & 95.4 & 55.9 & 90.3 \\ 
    & CellDivision \cite{malmsten2021automated} & Neu.Com.\&App`21 &InceptionV3 & 97.4 & 91.5 & 62.4 & 93.9 & 93.6 & 89.4  & 69.7 & 95.9 \\
    & ESOD \cite{lockhart2021automating} & MICCAI`21 &VGG-16 & 97.0 & 92.6  & \underline{75.3} & \textbf{97.1} & \textbf{95.9} & 94.0 & 73.1 & \textbf{96.4}\\ 
    \midrule
     \multirow{4}{*}{\rotatebox{90}{Ours}} &  CNNs (baseline) & & ResNet-50  & 98.9 & \underline{93.4} & 62.2  & 96.0  & 94.1  & 89.2  & \underline{78.9} & 94.7 \\
    & CNNs-Trans & & ResNet-50  & \underline{99.3} & 91.4 & 65.6  & 87.7  & 93.9  & \textbf{97.1} & 50.5 & 83.1 \\
    & CNNs-Trans-CRF & & ResNet-50  & 98.1  & 92.5  & 75.1 & 88.9 & 93.6 & 95.7 & 66.3 & 94.6 \\
    & \textbf{EmbryosFormer} & & ResNet-50 &  \textbf{99.7} & \textbf{94.7} & \textbf{80.6} & \underline{96.7} & \underline{94.4} & \underline{96.4} & \textbf{82.7} & \underline{96.1} \\
    \bottomrule
    \end{tabular}}
    \label{tab:accuracy_human}
\end{table*}

\begin{table}[!ht]
    \centering
    \caption{Comparison of average inference time on each time-lapse video. The scores are reported in second.}
    \resizebox{0.9\linewidth}{!}{
    \begin{tabular}{c|ll|c}
    \toprule
     & \textbf{Method} & \textbf{Feature} & \textbf{ Inf.time $\downarrow$} \\
     \toprule
     & CNNs-CRF \cite{lukyanenko2021developmental} & ResNet-50 & 2.26 \\ 
     & CellDivision \cite{malmsten2021automated} & InceptionV3 & 2.58  \\
     & ESOD \cite{lockhart2021automating} & VGG-16 & 2.27 \\ 
     \midrule 
     \multirow{4}{*}{\rotatebox{90}{Ours}} & CNNs (baseline)& ResNet-50  & 1.76 \\
     & CNNs-Trans & ResNet-50 & 2.06 \\
     & CNNs-Trans-CRF & ResNet-50 & 2.57 \\
     & \textbf{EmbryosFormer} & ResNet-50 & 1.79 \\ 
    \bottomrule
    \end{tabular}}
    \label{tab:abl_computation}
\end{table}

We further analyse the effectiveness of our proposed EmbryosFormer by two following ablation studies. First, we compare network computation complexity as shown in Table \ref{tab:abl_computation}, then we investigate the impact of each head as shown in Table \ref{tab:abl_head}. 

In the first ablation study, we compare inference time between our EmbryosFormer with others during inference. Table \ref{tab:abl_computation} reports the average of per-video time consumption during testing  on both datasets. The CNNs-CRF \cite{lukyanenko2021developmental} offers an efficient way to embed the monotonic constraint into the learning process. However, it takes a long time to compute e entire input sequence. Using InceptionV3 as the backbone for feature extraction, CellDivision \cite{malmsten2021automated} obtains potential results. However, it suffers from inadequate training and inference time consumption. Using ResNet-50 as a backbone network for visual encoding, EmbryosFormer consumes only 1.79s (seconds) to predict embryos stage development of a time-lapse video consisting of more than 300 images (314 images on Embryos Mouse and 339 images on Embryos Human, respectively). Compared to the ResNet-50 baseline, our EmbryosFormer inference time is compatible with only 0.03s difference.  The proposed method is not only efficient at performance but also at inference time, which is less than 0.47s, 0.79s, 0.48s compared to the existing works \cite{lukyanenko2021developmental, lockhart2021automating, malmsten2021automated}.


In the second ablation study, we compare the performance of global accuracy, the average of per-class precision and per-class recall with the following experiments. 
\noindent
 $\bullet$ Exp\#1: Classification Head - We remove the decoder branch and only use the encoder outputs to perform image-wise classification.
 
\noindent
$\bullet$  Exp\#2: Segmentation Head - We remove the Classification head in the encoder branch and use the segmentation output as the final prediction.
 
\noindent
$\bullet$  Exp\#3: Classification Head \& Segmentation Head - We setup as same as Exp\#2 but with the Classification head.
  
\noindent
$\bullet$  Exp\#4: Classification Head \& Segmentation Head \& Collaboration Head - This is our proposed EmbryosFormer, which aligns the coarse prediction from the classification head with monotonic non-decreasing constraint from the segmentation head and incorporates them into the collaboration head. 

The ablation study on the performance comparison between experiments is shown in Table \ref{tab:abl_head}. From this table, Exps\#2, \#3 demonstrate the  effectiveness of our design on the segmentation head. Without the need for any post-processing methods, we effectively produce sequence predictions and achieve better performance than off-the-shelf approaches (Table \ref{tab:global}, Table \ref{tab:abl_computation}) in an adequate time of processing.
Furthermore, the collaboration head (Exp\#4) improves the image-wise prediction by 0.7\% on global accuracy, 0.4\% and 1.6\% for average precision and recall, respectively, compared to the classification head alone (Exp\#1).

\begin{table}[thb]
\caption{Ablation studies on the effectiveness of EmbryosFormer with various settings. Classification, Segmentation, and Collaboration are denoted as Cls., Seg., and Col.}
\begin{tabular}{l|ccc|ccc}
\toprule
    \multirow{2}{*}{\textbf{Exp}} & \multicolumn{3}{c}{\textbf{Heads}}  & \multicolumn{3}{c}{\textbf{Performance}} \\ \cline{2-7}
 & Cls.  & Seg. & Col. &\shortstack{Global\\Acc.$\uparrow$} & \shortstack{Avg\\Prec.$\uparrow$} & \shortstack{vg\\Rec.$\uparrow$} \\ \toprule
  \#1  & \checkmark &  \xmark  & \xmark &   93.6  & 92.5 & 90.8 \\
 \#2   & \xmark  & \checkmark & \xmark &  93.2  & 92.1 & 89.9 \\
 \#3   &  \checkmark  & \checkmark   & \xmark &  93.6 & 92.3 & 90.6 \\
\midrule
 \#4   &  \checkmark  &  \checkmark  & \checkmark &  94.3 & 92.9 & 92.4 \\\bottomrule
\end{tabular}
 \label{tab:abl_head}
\end{table}


\subsection*{Conclusion \& Discussion}
This paper presents EmbryosFormer, an embryo stage development classification framework with collaborative heads encoding-decoding. The proposed EmbryosFormer is implemented based on the deformable transformer with three heads. The classification head aims to predict the per-image label without taking monotonic non-decreasing constraint into consideration, which is then handled by the segmentation head. The collaboration head incorporates per-image prediction from the classification head and per-class prediction from the segmentation head. Experiments and ablation studies on two benchmark datasets of both Embryos Mouse and Embryos Human show that EmbryosFormer can accurately predict embryo stage development and surpass state-of-the-art methods. 

Future investigations might aim for better techniques to work on original images directly collected from an incubator system. Techniques for spatial attention such as \cite{cordonnier2021differentiable, li2022contextual} and self-supervised learning \cite{chen2020simple, caron2021emerging} are also potential extensions for performance improvement. 

\section*{Acknowledgment}
This material is based upon work supported by the National Science Foundation under Award No OIA-1946391 and NSF 1920920. 
Tien-Phat Nguyen was funded by Vingroup JSC, Vingroup Innovation Foundation (VINIF), Institute of Big Data, VINIF.2021.ThS.JVN.04.
Trong-Thang Pham was funded by Vingroup JSC, Vingroup Innovation Foundation (VINIF), Institute of Big Data, VINIF.2021.ThS.JVN.05.
Minh-Triet Tran was funded by Vingroup and supported by Vingroup Innovation Foundation (VINIF) under project code VINIF.2019.DA19.
\newpage
{\small
\bibliographystyle{ieee_fullname}
\bibliography{egbib}
}

\end{document}


\title{EmbryosFormer: Deformable Transformer and Collaborative Encoding-Decoding for Embryos Stage Development Classification \\
\textit{Supplementary Material}}

\author{Tien-Phat Nguyen\textsuperscript{1,4}, Trong-Thang Pham\textsuperscript{*4}, Tri Nguyen*\textsuperscript{*7}, Hieu Le\textsuperscript{*1}, Dung Nguyen\textsuperscript{5}, \\ Hau Lam\textsuperscript{6}, Phong Nguyen\textsuperscript{1}, Jennifer Fowler\textsuperscript{8}, Minh-Triet Tran\textsuperscript{2,3,4}, Ngan Le\textsuperscript{9} \\
\small \textsuperscript{1}FPT Software AI Center, Ho Chi Minh City, Vietnam \\
\small \textsuperscript{2}University of Science, VNU-HCM; 
\small \textsuperscript{3}Vietnam National University, Ho Chi Minh City, Vietnam\\
\small \textsuperscript{4}John von Neumann Institute, Vietnam National University, Ho Chi Minh City, Vietnam\\
\small \textsuperscript{5}IVFMD, My Duc Phu Nhuan hospital, Ho Chi Minh City, Vietnam\\
\small \textsuperscript{6}Olea Fertility, Vinmec Central Park International Hospital, Ho Chi Minh City, Vietnam\\
\small \textsuperscript{7}HOPE Research Center, My Duc Hospital, Ho Chi Minh City, Vietnam \\
\small \textsuperscript{8}Arkansas Economic Development Commission, Little Rock, AR USA 72202\\
\small \textsuperscript{9}Department of Computer Science and Computer Engineering, University of Arkansas, Fayetteville, AR, USA 72703\\
\small \textsuperscript{*}\textit{equal contribution}\\
}

\maketitle

Our EmbryosFormer network is optimized by collaboratively learning three heads corresponding to the classification head, segmentation head and collaboration head. While the classification head is a per-image classification and it has been commonly used in most existing embryo stage development approaches; it can not handle the monotonic non-decreasing constraint. We proposed a segmentation head at the deformable transformer decoder path to ensure temporal coherence with the monotonic constraint. Furthermore, we proposed a collaboration head which collaboratively takes the merits of both classification head and segmentation heads. All three heads are trained in an end-to-end manner. 

In this supplementary, we will first analyse the effectiveness of our proposed segmentation head and collaboration head in handling the monotonic non-decreasing constraint. We will then provide a qualitative comparison between our EmbryosFormer with other state-of-the-art methods. As shown in Table \ref{tb:stage-level}, the Embryos Mouse dataset contains more cell stages development, while the number of images belonging to each stage is more imbalanced compared to Embryos Human dataset. Thus, we choose Embryos Human dataset to conduct the qualitative analysis in this supplementary.

\begin{table*}[ht!]
\centering
\caption{Stage-level data distribution of each cell stage in Embryos Mouse and Embryos Human datasets.}
\resizebox{0.9\textwidth}{!}{
\begin{tabular}{l|rrrrrrrr|rrrr}
\toprule
      \textbf{Images/} & \multicolumn{8}{c|}{\textbf{Mouse}} &  \multicolumn{4}{|c}{\textbf{Human}}  \\ \cline{2-13} 
      \textbf{Stage} & 1     & 2 & 3 & 4 & 5 & 6 & 7 & 8 & 1     & 2 & 3 & 4 \\ \toprule
Average  &  50.02  & 165.32  & 8.28  & 88.2  &  4.81  & 8.05 & 8.13 & 2  & 63.6 & 164.7 & 41.3 & 68.8 \\ 
\bottomrule
\end{tabular}
}
\label{tb:stage-level}
\end{table*}
\section*{Dataset Statistics}
In this section, we provide the stage-level statistical information of both the Embryos Mouse and Embryos Human datasets. The stage-level level is computed as the distribution of the number of images belonging to each cell stage development.

\section*{Classification head v.s segmentation head}
\begin{figure*}[h!]
\centering
    \subfigure{
        \includegraphics[width=0.9\linewidth]{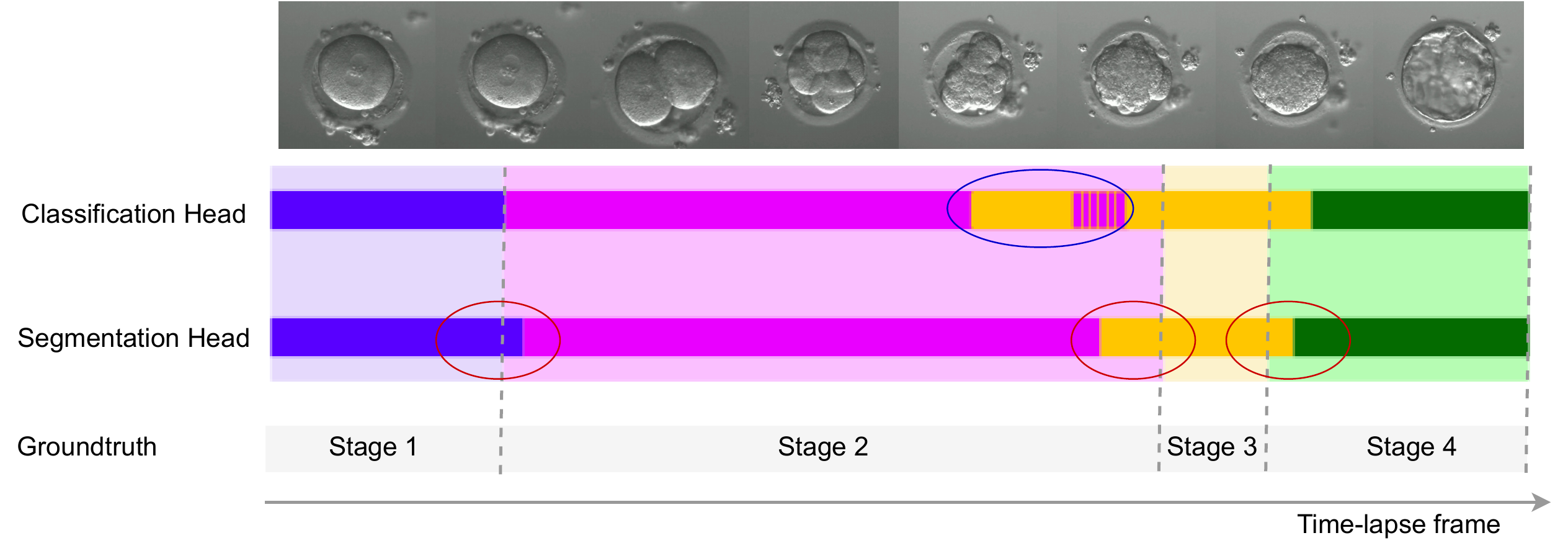}
    }
    \quad
    \subfigure{
        \includegraphics[width=0.9\linewidth]{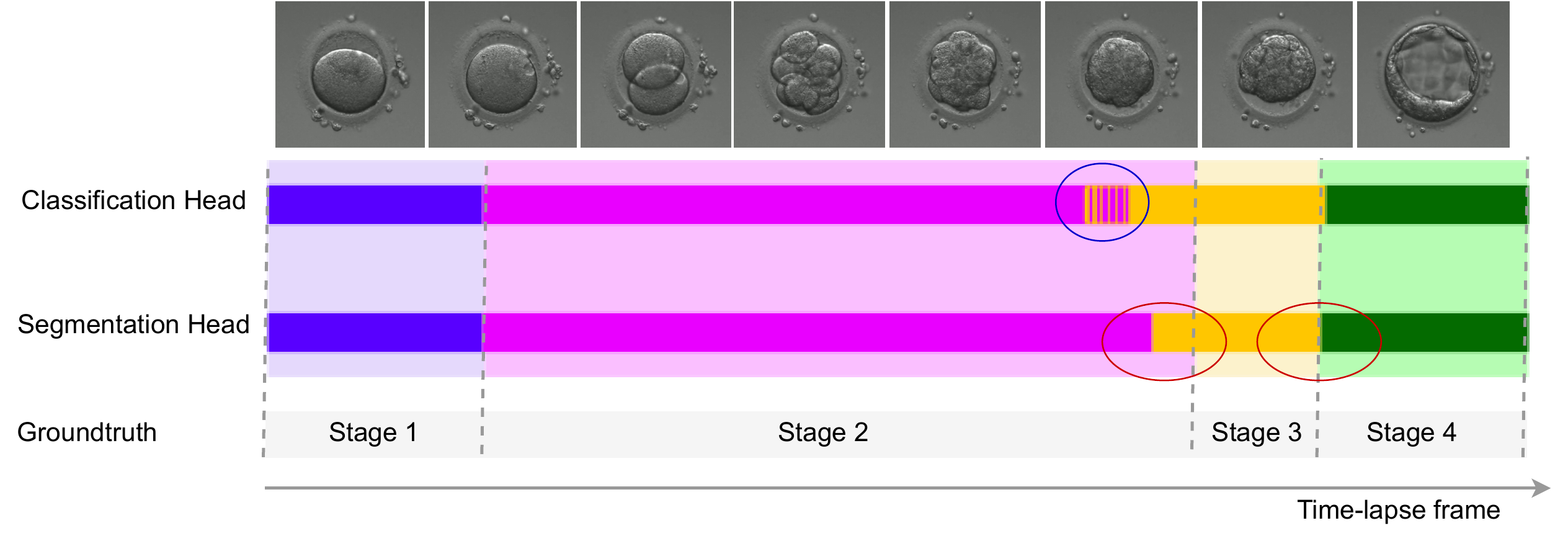}
    }
    \caption{Qualitative comparison between classification head and segmentation head on Human Embryos dataset with 4 cell development stages. There are two time-lapse embryo examples in this figure. In those examples, the miss-classified within-stage images by classification head are highlighted in \textcolor{blue}{blue circle} and miss-classified transitional images as well as within-stage classification improvement by segmentation head are highlighted in \textcolor{red}{red circle}.}
    \label{fig:cls_seg}
\end{figure*}

As shown in Fig.\ref{fig:cls_seg}, the classification head, which focuses on image-level classification, generates promising results. However, it does not include monotonic constraint and thus, it likely miss-classifies within-stage images as shown in \textcolor{blue}{blue circle}. On the other hand, the segmentation head is able to handle temporal coherence with monotonic constraint, and thus, the within-stage miss-classification problem is eliminated. However, the segmentation head is still limited at classifying transitional images, as shown in \textcolor{red}{red circle}.

\begin{figure*}[h!]
\centering
    \subfigure{
        \includegraphics[width=0.9\linewidth]{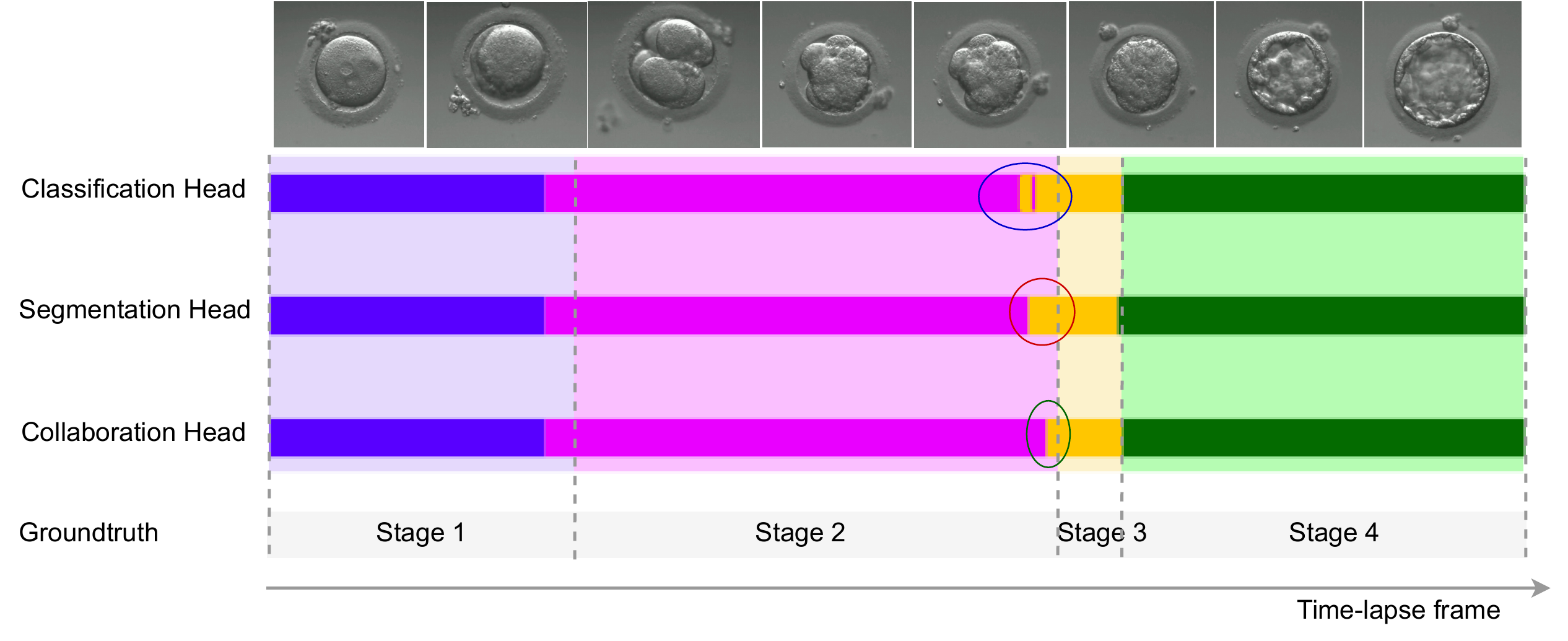}
    }
    \quad
    \subfigure{
        \includegraphics[width=0.9\linewidth]{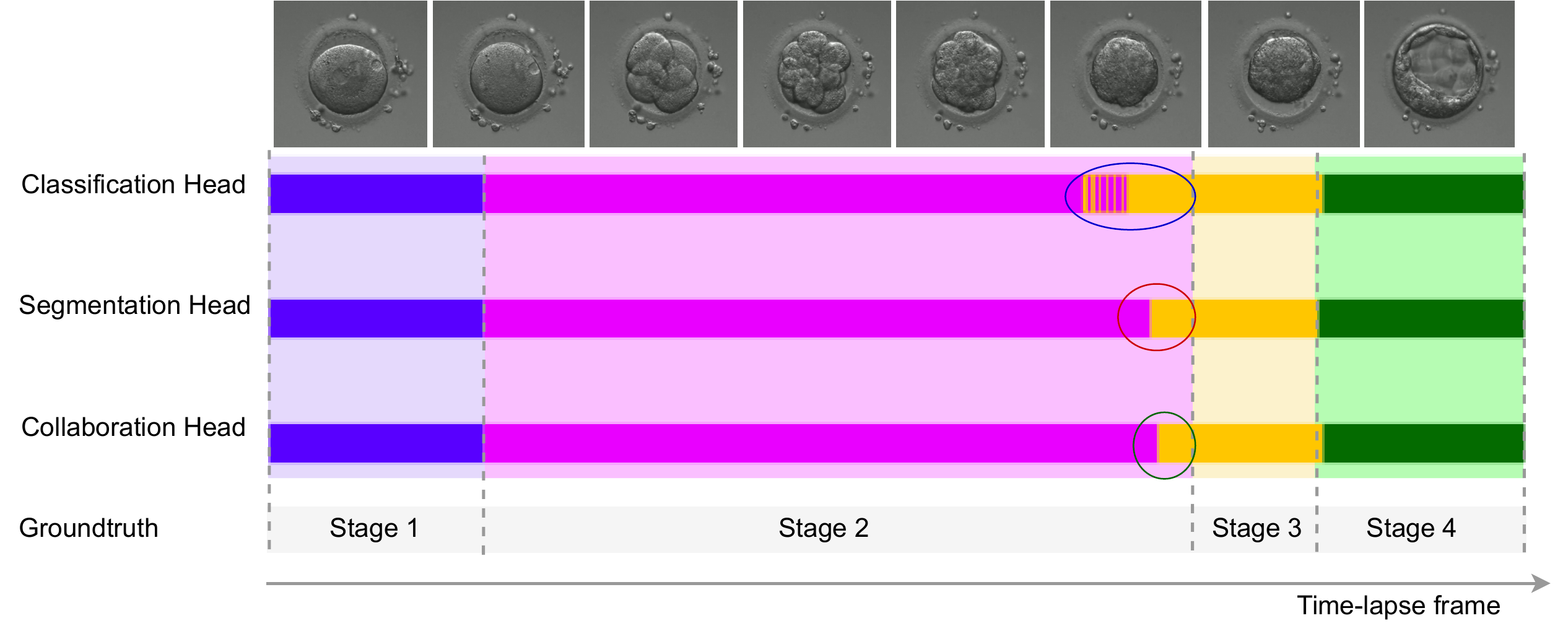}
    }
    \caption{Qualitative comparison between classification head, segmentation head and collaboration on Human Embryos dataset with 4 cell developmental stages. There are two time-lapse embryo examples in this figure. In those examples, the miss-classified within-stage images by classification head are highlighted in \textcolor{blue}{blue circle} and miss-classified transitional images as well as within-stage classification improvement by segmentation head are highlighted in \textcolor{red}{red circle}. The within-stage classification improvement and transitional images classification improvement by the collaboration head is shown in \textcolor{green}{green circle}.}
    \label{fig:cls_seg_col}
\end{figure*}

\section*{Classification head v.s segmentation head v.s collaboration head}
As shown in the Fig.\ref{fig:cls_seg_col}, while the classification head miss-classifies within-stage images \textcolor{blue}{blue circle} because monotonic constraint is not included, segmentation head handles monotonic constraint but miss-classifies transitional images \textcolor{red}{red circle}. We proposed a collaboration head to incorporate the advantages of both the classification head and segmentation head. The improvement by the collaboration head is shown in \textcolor{green}{green circle} which qualitatively and quantitatively (as in the main manuscript) obtains better performance compared to individual classification and segmentation heads.



\section*{Qualitative comparisons}
\begin{figure*}[h!]
\centering
    \subfigure{
        \includegraphics[width=0.9\linewidth]{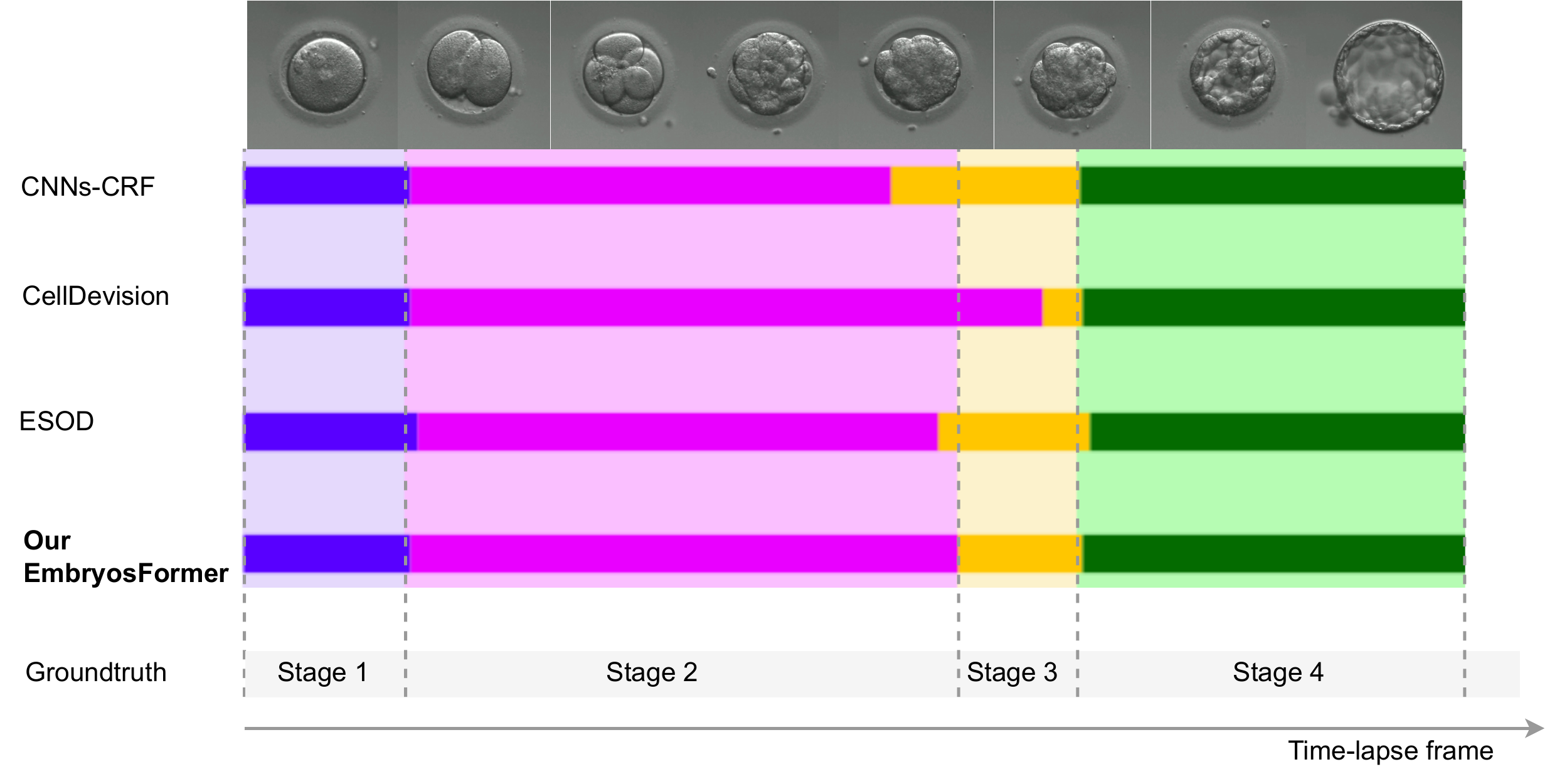}
    }
    \quad
    \subfigure{
        \includegraphics[width=0.9\linewidth]{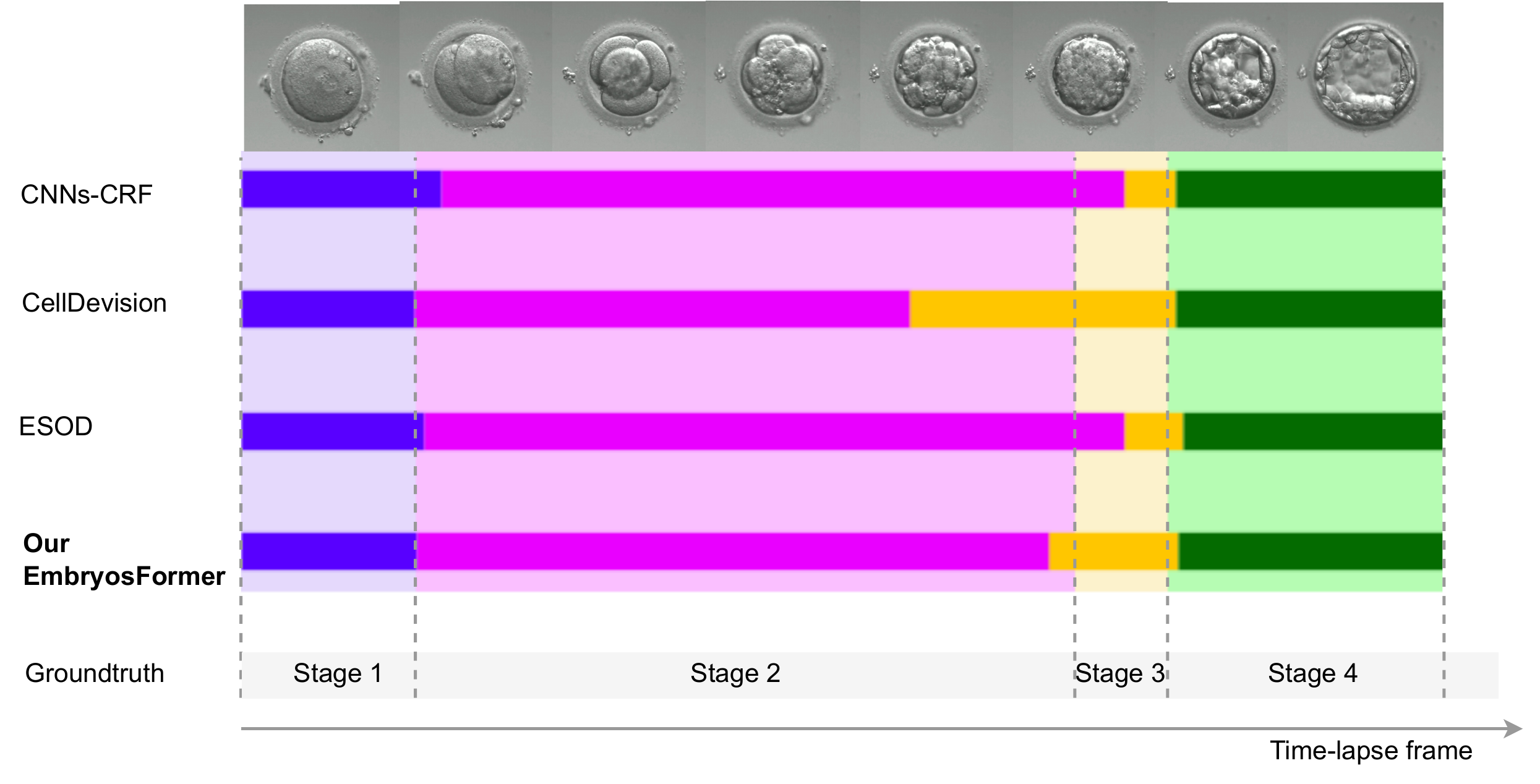}
    }
    \caption{Visualization of qualitative comparison between CNNs-CRF~\cite{lukyanenko2021developmental} ($2^{nd}$ row), CellDivision~\cite{malmsten2021automated} ($3^{rd}$ row), and ESOD \cite{lockhart2021automating} ($4^{th}$ row) with our EmbryosFormer ($5^{th}$ row) on Human Embryos dataset with two time-lapse embryos examples.}
    \label{fig:compare}
\end{figure*}

In this section, we provide some qualitative comparisons between our EmbryosFormer and the existing state-of-the-art cell stage development. As shown in Fig.\ref{fig:compare}, our proposed EmbryosFormer with three heads is able to archive better performance at both within-stage and transition images compared to CNNs-CRF~\cite{lukyanenko2021developmental}, CellDivision~\cite{malmsten2021automated}, and ESOD \cite{lockhart2021automating}.

\newpage
{\small
\bibliographystyle{ieee_fullname}
\bibliography{egbib}
}